\documentclass[9pt]{IEEEtran}

\newcommand{\subparagraph}{}
\usepackage[compact]{titlesec}

\usepackage{graphicx}
\usepackage{amsmath}
\usepackage{amssymb}
\usepackage{algorithmic}
\usepackage{subcaption}
\usepackage{stfloats}
\usepackage{float}
\usepackage{url}
\usepackage[font = scriptsize]{caption}
\usepackage{xcolor}
\usepackage[linesnumbered,ruled,vlined]{algorithm2e}
\usepackage{fancyhdr}

\SetCommentSty{mycommfont}

\begin{document}

\bstctlcite{IEEEexample:BSTcontrol}

\title{Security Analysis and Enhancement of Model Compressed Deep Learning Systems under Adversarial Attacks}

\author{
\IEEEauthorblockN{Qi Liu\IEEEauthorrefmark{1}, Tao Liu\IEEEauthorrefmark{1}, Zihao Liu\IEEEauthorrefmark{1}, Yanzhi Wang\IEEEauthorrefmark{2}, Yier Jin\IEEEauthorrefmark{3} and Wujie Wen\IEEEauthorrefmark{1}}

\IEEEauthorblockA{\IEEEauthorrefmark{1}\textit{Florida International University}, \IEEEauthorrefmark{2}\textit{Syracuse University}, 
\IEEEauthorrefmark{3}\textit{University of Florida}\\
\IEEEauthorrefmark{1}\{qliu020, tliu023, zliu021, wwen\}@fiu.edu, \IEEEauthorrefmark{2}ywang393@syr.edu, \IEEEauthorrefmark{3}yier.jin@ece.ufl.edu}
\vspace{-25pt}

\thanks{This work is supported by the 2016-2017 Collaborative Seed Award Program of Florida Center for Cybersecurity (FC$^2$).}

%2017 VFRP from US ARFL/RI, Rome, NY. Any opinions, findings, and conclusions or recommendations expressed in this material are those of the authors and do not necessarily reflect the views of AFRL or its contractors.}

}

\maketitle

\thispagestyle{fancy}
\lhead{}\chead{}\rhead{}
\lfoot{}\cfoot{\thepage}\rfoot{}
%\cfoot{\thepage}
\renewcommand{\headrulewidth}{0pt}
\renewcommand{\footrulewidth}{0pt}
\pagestyle{fancy}

\begin{abstract}
Thanks to recent machine learning model innovation and computing hardware advancement, the state-of-the-art of Deep Neural Network (DNN) is presenting human-level performance for many complex intelligent tasks in real-world applications.
%along with machine learning model innovation and recent hardware advancement.
%The recent impressive hardware advancement is demonstrating unsurpassed acceleration of deploying, 
However, it also introduces ever-increasing security concerns for those intelligent systems. For example, the emerging adversarial attacks indicate that even very small and often imperceptible adversarial input perturbations can easily mislead the cognitive function of deep learning systems (DLS). 
%Despite of many relevant researches on DNN security, 
Existing DNN adversarial studies are narrowly performed on the ideal software-level DNN models with a focus on single uncertainty factor, i.e. input perturbations, however, the impact of DNN model reshaping on adversarial attacks, which is introduced by various hardware-favorable techniques such as hash-based weight compression during modern DNN hardware implementation, has never been discussed.
%based on the single uncertainty factor, i.e. input perturbations, with a focus on the ideal software-level DNN models.
%and are solely performed on ideal software-level DNN models without considering DNN model reshaping introduced by various hardware-favorable techniques such as network pruning and HashNet during practical task executions in DNN hardware platforms. 
%The questions like how those software-based attacks can be exerted in practical DNN models and during realistic task execution are of critical importance but remain unexplored. 
In this work, we for the first time investigate the multi-factor adversarial attack problem in practical model optimized deep learning systems by jointly considering the DNN model-reshaping (e.g. HashNet based deep compression) and the input perturbations. We first augment adversarial example generating method dedicated to the compressed DNN models by incorporating the software-based approaches and mathematical modeled DNN reshaping. 
%in the DNN model-reshaped practical deep learning systems by integrating highly compressed weights with input perturbations.
%the reshaped model has been characterized to develop the multi-factors adversarial attacks in deep learning systems by integrating highly compressed weights with input perturbations. 
We then conduct a comprehensive robustness and vulnerability analysis of deep compressed DNN models under derived adversarial attacks. %the proposed multi-factors adversarial attacks. 
%Inspired from the security analysis, 
A defense technique named ``gradient inhibition" is further developed to ease the generating of adversarial examples thus to effectively mitigate adversarial attacks towards both software and hardware-oriented DNNs. Simulation results show that ``gradient inhibition" can decrease the average success rate of adversarial attacks from 87.99\% to 4.77\% (from 86.74\% to 4.64\%) on MNIST (CIFAR-10) benchmark with marginal accuracy degradation across various DNNs.

%\textcolor{red}{Simulation results well validate our proposed solution: Update after result part}

%However, the expensive computation and massive storage requirement significantly hinder its applications on resource limited platform. 
%The hardware-oriented Deep Learning System (DLS) become a promising solution for low-power, high-speed data processing greatly enriches intelligent applications. 

%Recently, security issues like adversarial attacks draw increasing attentions in DNN researches. Even inappreciable small input perturbations may severely destroy the cognitive functionality. 

%Unfortunately, the insidious security concerns have never been addressed in hardware designs, exposing serious threats to DLSs, thus urgently requirements of security analysis on vulnerabilities. However, all the existing DNN adversarial studies are based on the single uncertainty factor, i.e. input perturbations, and are solely performed on ideal software-level DNN models without considering the unique model reshaping during practical executions in DLS.

%Whether and how those software-based attacks and defense solutions can be exerted and/or exploited in practical DLSs remain unexplored.% and deserve thorough investigations.

%Our experimental results show that by adopting ``gradient inhibition'', the success rate of adversarial attacks has been reduced remarkably from 95.81\% to 0.22\% on MNIST benchmark.
\end{abstract}

\section{Introduction}

As one of the most fascinating techniques when we are entering the era of Artificial Intelligent (AI), Deep Neural Networks (DNNs) are penetrating the real world in many exciting applications such as image processing, face recognition, self-driving cars, robotics and machine translations etc. 
%As one of the latest breaking news of AI, \textit{``AlphaGo''}, the deep learning system developed by Google-owned company ``DeepMind'', has again steamrolled the world's number one ``Go'' player into a 3-0 defeat in ``Go'' game after crushing both European and World champions last year. This further highlights the remarkable capability of DNNs for handling extremely complicated tasks~\cite{silver2016mastering,web_go_kejie}.
Nonetheless, all this success, to great extent, is enabled by introducing the powerful data analysis capability of state-of-the-art large-scale DNNs with deep and complex structures and huge volume of model parameters, significantly exacerbating the demand for computing resource and data storage of hardware platforms.  As an example, the large-scale image classification implementation of famous deep convolutional neural network (CNN) ``AlextNet'' involves 61 million parameters off-chip memory accesses and 1.5 billion high precision floating-point operations~\cite{krizhevsky2012imagenet}. 

Fortunately, recent hardware engine innovation enables the implementations of those once ``conceptual'' DNN software systems in both high-performance computing and resource-limited embedded platforms for performing various intelligent tasks~\cite{web5, web6, web7}. Many hardware-favorable DNN architectures along with various DNN model optimization techniques are developed to accelerate dedicated computations on general-purpose platforms like GPU~\cite{ciresan2011flexible} and CPU~\cite{vanhoucke2011improving}, domain-specific hardware like FPGA~\cite{farabet2011large}, and customized ASIC, e.g. recent Google Tensor Processing Unit (TPU)~\cite{web4,jouppi2017datacenter,web_TPU}.

%have been widely used in many exciting applications across image processing, face recognition, self-driving cars, robotics and machine translations etc. 
While DNN's broad and positive impacts along with its impressive hardware advancement excite multiple industries in myriad ways, it also brings about ever-increasing security challenges. Since the classification results of DNN systems are usually derived from the probabilities~\cite{ghahramani2015probabilistic,barreno2010security}, the attackers can easily compromise system security by exploiting specific vulnerabilities of learning algorithms or classifiers through a careful manipulation of the input data samples, namely \textit{adversarial examples}, i.e. circumvent anomaly detection, misclassify the adversarial images at testing time~\cite{goodfellow2014explaining} or adversarially manipulate the perceptual systems of autonomous vehicles to the misreading of road signs, thus causing potential disastrous consequences~\cite{papernot2016practical}. Hence, safeguarding the security of DNN systems has become an urgent task. 

Many DNN adversarial researches have been conducted, including adversarial example generating~\cite{goodfellow2014explaining,papernot2016limitations}, robustness analysis~\cite{fawzi2015analysis} and mitigation techniques~\cite{goodfellow2014explaining,gu2014towards,papernot2016distillation}. However, existing adversarial studies focus only on software-level DNN models by (over-) simply assuming that the input perturbations are the only uncertain factor under unchanged software-level DNN models. The additional DNN model change, e.g. non-linear weights reshaping to largely compress DNN scale~\cite{chen2015compressing,cao2017hashnet,han2015learning,han2015deep}, which is inevitable because of the hardware resource constraints during DNN deployment, is often neglected.   
%The hardware constraints of DNN deployment inevitably introduces many hardware-oriented techniques to dramatically reshape DNN models, i.e. non-linear weights transformation to compress DNN scale~\cite{chen2015compressing,cao2017hashnet,han2015learning,han2015deep} 
As we shall present in section~\ref{attack_design}, the adversarial attack to practical DNN systems will be a multi-factor problem rather than the ideal single-factor problem from crafted adversarial inputs. Since the realistic tasks often need to be executed in DNN hardware systems with extra efforts on model compression, discovering the nature of more realistic adversarial attacks, as well as developing effective countermeasures to protect such practical learning systems will be of critical importance at the early stage of DNN applications.

In this work, we \textbf{for the first time formulate the multi-factor adversarial attacks} tailored for the practical deep learning systems by integrating the mathematical modeled DNN model reshaping (take hash-based DNN weight compression as an example) and the input perturbations. We then \textbf{for the first attempt to systematically analyze the interplays among the hash compression ratio, the amplitude of input perturbations, adversarial attack successful rate and accuracy} through extensive experimental and theoretical studies. Interestingly, we discover that the hash-based deep compressed DNN models can be somewhat less vulnerable to adversarial attacks because of the reshaped weight distribution when compared to the uncompressed software-DNN models. Inspired by this observation, a defense technique named ``gradient inhibition" is further proposed to suppress the generating of input perturbations thus to effectively prevent the adversarial attacks for deep learning systems. Experimental results show that ``gradient inhibition" can reduce the success rate of adversarial attacks from 87.99\% to 4.77\% (from 86.74\% to 4.64\%) on average on MNIST~\cite{lecun1998mnist} (CIFAR-10~\cite{krizhevsky2009learning}) benchmark %with marginal accuracy degradation across various DNNs.
while maintaining the same level of accuracy across various DNNs.

\section{Preliminary}

\subsection{Basics of Deep Neural Networks}
Deep Neural Network (DNN) introduces multiple layers with complex structures to model a high-level abstraction of the data~\cite{hinton2006reducing}, and exhibits high effectiveness in cognitive applications by leveraging the deep cascaded layer structure~\cite{krizhevsky2012imagenet,simonyan2014very,szegedy2015going}. %Figure~\ref{fig:cnn} shows 
For example, a typical modern DNN often consists of following types of layers: The convolutional layer extracts sufficient feature maps from the last layer by applying kernel-based convolutions, the pooling layer performs a downsampling operation (max or average pooling) along the spatial dimensions for a volume reduction, and the fully-connected layer further computes the class score based on the final weighted results and the non-linear activation functions.

%As modern DNNs become more powerful with an ever-increasing model size, i.e. 60M to even 10B parameters to represent the weight connections~\cite{cheng2015exploration,han2016eie, han2015deep}, reducing their storage and computational costs becomes critical to meet the requirements of practical applications in hardware oriented Deep Learning Systems with limited resources, i.e. ASIC or FPGA. %Recent works show that fetching large number of off-chip weights from the external DRAM is even two to three orders of magnitude more expensive than ALU operations, i.e. 12.8 Watts for just DRAM access when testing a DNN with 1G connections at 20Hz~\cite{han2016eie}, far beyond the power envelope of many devices. 
%Therefore, removing the redundancy of DNN models has become a ``must-have" step in Deep Learning System design~\cite{han2015deep}. 

%\begin{figure}[t]
%\centering
%\includegraphics[width = 1\columnwidth]{figs/CNN_structure}
%\caption{A conceptual view of the typical structure of state-of-the-art CNN}
%\label{fig:cnn}
%\end{figure}

\subsection{Model Reshaping}
As modern DNNs become more powerful with an ever-increasing model size, i.e. 60M to even 10B parameters to represent the weight connections~\cite{cheng2015exploration,han2016eie, han2015deep}, reducing their storage and computational costs becomes critical to meet the requirement of practical applications in hardware-oriented DNNs with limited resources, i.e. ASIC or FPGA. 
Therefore, removing the redundancy of DNN models has become a ``must-have" step in deep learning system design~\cite{han2015deep}.

\begin{figure}[t]
\centering
\includegraphics[width = 0.87\columnwidth]{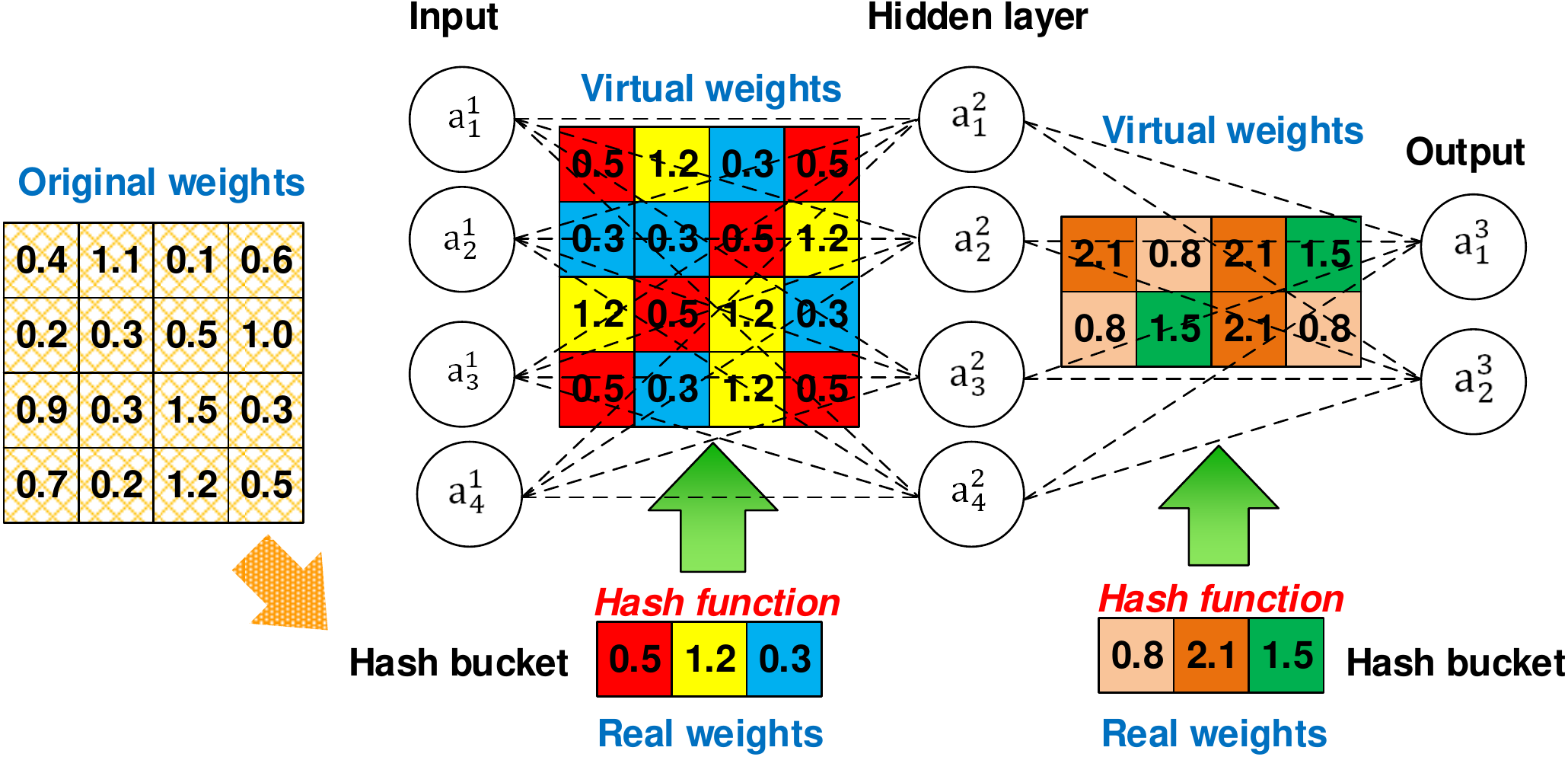}
\caption{Illustration of model reshaping in an example HashNet.}
\label{fig:Hash}
\vspace{-10pt}
\end{figure}

Many studies are preformed to reshape the DNN models towards affordable hardware implementations, including network pruning~\cite{han2015learning,han2015deep}, HashNet~\cite{chen2015compressing,cao2017hashnet}, etc. Those solutions can effectively compress the weights through some non-linear transformations. Take the HashNet adopted in this work as an example, a hash function is selected to randomly group connection weights into hash buckets. All connections within the same hash bucket share a single parameter value. Therefore, the needed memory to store the weights can be significantly reduced. Figure~\ref{fig:Hash} demonstrates the idea of an example HashNet for achieving significant storage reduction with limited accuracy loss. The original weights are converted to real weights by a random hash procedure. The real weights together with the hash index which are physically stored in the DNN hardware only cost $\sim20\%$ memory space compared to that of original (virtual) weights. 

\subsection{Adversarial Examples}
Adversarial examples are maliciously crafted inputs dedicated to mislead the DNN classification by introducing small input perturbations. The generating of adversarial examples can be modeled as an optimization problem:
\begin{equation}
\arg \min_{\delta_X}\parallel\delta_X\parallel\ s.t.\ F\left( X+\delta_X \right) = Y^*
\label{AE_Model} 
\end{equation}
Here $F$ represents the function of target DNN model, and is usually determined by the detailed DNN configurations such as the architecture and the weight. $Y^*$ is the distorted output which is different from the correct output $Y$. $X^*=X+\delta_X$ denotes the adversarial example perturbed by $\delta_X$. Hence the question becomes how to solve the optimization problem to find the minimized $\delta_X$. The common approach to derive adversarial examples is to extract adversarial perturbations $\delta_X$ from the gradient information, since the gradient is a good measurement for the output response difference with respect to variations introduced in each dimension of an input vector. Hence, there are two gradient-based methods to generate adversarial examples from software DNN models: \textit{Fast Gradient Sign Method} (FGSM)~\cite{goodfellow2014explaining} and \textit{Jacobian-based Saliency Map Approach} (JSMA)~\cite{papernot2016limitations}. The former adds a small perturbation in the direction of the sign of the gradient of the loss function with respect to the input of the DNN to all input dimensions, while the latter only distorts the most significant input features based on the salience map extracted from gradient of model function w.r.t. inputs--Jacobian matrix.  
%The common practice of these two methods is to extract 
%are two major approaches to generate the adversarial examples. 

%\textcolor{red}{Dr. Wen: We need to at least introduce what is FGSM, and what is JSMA, otherwise, people do NOT know especially for reviewers not in this field}
%\subsubsection{Fast Gradient Sign Method(FGSM)}
%This method add a small perturbation in the direction of the sign of the gradient of the loss function with respect to the input of the DNN to all input dimensions.
 %\begin{equation}
%x^*=x+\epsilon sign(\nabla_{x} J( \theta ,x,y))
% \end{equation}
%\subsubsection{Jacobian-based Saliency Map Approach(JSMA)}
%This method computes the Jacobian
%matrix of the model F and ranks features in a saliency map to achieve the adversarial goal.

\begin{figure}[t]
\centering
\includegraphics[width = 0.98\columnwidth]{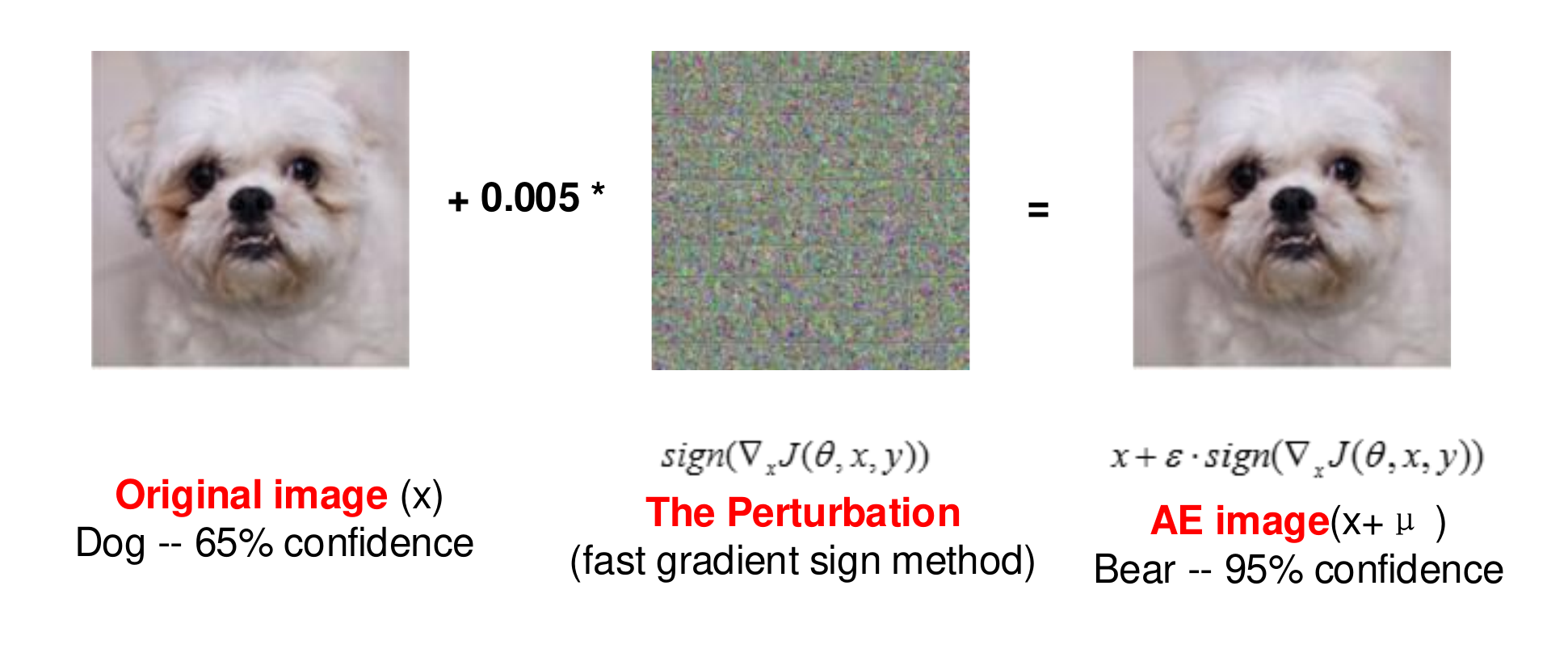}
\caption{Illustration of adversarial examples}
\label{fig:AE}
\vspace{-10pt}
\end{figure}

Figure~\ref{fig:AE} shows a conceptual view of \textit{FGSM} based adversary on a representative DNN model--``AlexNet" with perturbation parameter $\epsilon=0.005$. The image originally correctly classified as ``Dog'' by the ``AlexNet" (65\% confidence) is now misclassified as ``Bear'' with a much higher confidence (95\%) due to the slightly polluted input. However, such an adversarial example is so close to the original image that the differences are indistinguishable to human eyes.

%\subsection{Limitations of Existing DNN Security Researches}

%Existing researches present a complete discussions on DNN security under adversarial attacks, including adversary generating~\cite{}, robustness analysis~\cite{} and mitigation techniques~\cite{}. 
%However, exerting adversarial attacks on Deep Learning System is non-trivial due to the specific hardware oriented model reshaping and limited processing capabilities. On one hand, though weights and gradients are fundamental in adversary generating, both FGSM and JSMA approaches are focus on the discussion of generating methodology through perpetuated inputs, but merely consider the weights pattern as a non-negligible factor that affecting the adversarial effectiveness, significantly hinder their application on Deep Learning System. On the other hand, the computational expensive robustness analysis and mitigation techniques are solely performed on software-level DNN models without considering the resource limitation and real-time threat defense in realistic application.

%In this work, 

% \subsubsection{Fast Gradient Sign Method (FGSM)}

% \subsubsection{Fast Gradient Sign Method (FGSM)}

\section{Attack Design}
\label{attack_design}
To analyze the vulnerabilities of practical deep learning systems under adversarial attacks, we first present the threat model, followed by an attack methodology developed for conducting adversarial attacks over the hash-based deep compressed DNN models. 
%Before introducing the attack design, we first present the threat model. 

\subsection{Threat Model}
In this work, we adopt a white-box adversarial attack model. We assume that the attacker has full access to all target compressed /non-compressed DNNs, training and testing dataset. The objective of adversarial attack is to mislead the classification of an original class to a different target, i.e. original $\neq$ target. To conduct the attack, the attacker first acquires the DNN model information such as weights, cost function, hash compression, gradient with normal input etc. Then the imperceptible perturbations are calculated through derived adversarial crafting algorithms and injected into normal inputs to generate adversarial examples. Finally, the adversarial examples will be sent to compressed/non-compressed DNN models, fooling the deep learning systems with adversarial classification results.

\subsection{Adversarial Attack Design}
To exert effective adversarial attacks to practical deep learning systems, our first step is to extend the single-factor adversarial examples generating algorithm to the multi-factor version based on 
%a multi-factor adversary generating algorithm is first proposed based on 
the augmentation of software-model oriented FGSM and JSMA approaches by taking 
mathematical characterized hash-based deep compression into consideration.
%the weights compression and model reshaping into consideration.  
Then a synthesized attack methodology is presented as our basis for security analysis and robustness evaluation.

\subsubsection{Multi-factor Adversarial Example Generating} 
%In our proposed two-factor adversary generating algorithm, 
To better illustrate how the adversary generating will be altered by the input perturbation and model reshaping in deep learning systems, the adversarial attack is again modeled as an optimization problem:
\begin{equation}
\arg \min_{\delta_{X_H}}\parallel\delta_{X_H}\parallel\ s.t.\ F_H\left( X+\delta_{X_H} \right) = Y^*
\label{AE_Hard_Model} 
\end{equation}
where $F_H$ represents the hardware-oriented hashed DNN model derived from its software version (or uncompressed DNN model) $F$ with marginal accuracy reduction. $Y^*$ is the distorted output which is different from the correct output $Y$. 
%$X^*=X+\delta_X$ denotes the adversarial example perturbed by $\delta_X$. 
%Hence the question becomes how to solve the optimization problem to find the minimized $\delta_X$. 
Apparently, the minimum input perturbations of $F_H$ ($\delta_{X_H}$) will be less likely to be equal to that of ideal software DNN model, i.e. $F$ ($\delta_{X}$), even for the same adversarial target $Y^*$: $min \delta_{X_H}\neq min \delta_{X}$ because of the model reshaping ($F_H \neq F$). 
%This indicates that the adversary examples generated from the uncompressed model $F$ may exhibit different effectiveness when applying to the hardware model $F_H$. 
If we define $W$ and $W_H=\varphi \left ( W \right)$ as the weight matrix of DNN model $F$ and $F_H$, the activation output will be $W\left(X + \delta_{X} \right)$ and $\varphi \left ( W \right) \left(X + \delta_{X_H} \right)$, respectively, where $\varphi \left ( W \right)$ denotes a hardware-oriented weight transformation--hashing in HashNet. Since the hardware-oriented model reshaping should always minimize the accuracy loss, the corresponding results after activation $f$ should be $f\left (WX \right) \approx f\left (\varphi \left ( W \right) X \right)$. However, the DNN output perturbations will be changed from $f\left (W\delta_{X} \right)$ to $f\left (\varphi \left ( W \right)\delta_{X_H} \right)$ accordingly. Even for the same adversarial example ($\delta_{X_H}= \delta_{X}$), the responses from the two models will be quite different. Different from the single uncertainty factor assumption, i.e. input perturbations, adopted in the software DNN models, the compressed version of adversarial attacks will be more complicated and become a multi-factor problem due to the additional weight transformations.

%Therefore, even for the same adversarial example ($\delta_{X_H}= \delta_{X}$), the responses from the two models will be quite different, 

%thus a two-factor problem for adversary generating in Deep Learning System.

%In this work, the HashNet is adopted to present the model reshaping in our adversarial attack design. 
 
As the foundation for hardware-oriented adversarial example generating, we first mathematically model the deep compressed DNN model--HashNet. In HashNet, the derived classification output $a_i^{l+1}$ for neuron $i$ in layer $l+1$ and the gradient ($\delta_j^l$) of loss function $\mathcal{L}$ over activation $j$ in layer $l$ can be presented as: 
%For the feed-forward pass, the classification output will be calculate as: 
\begin{equation}
a_i^{l+1}=f\left(\sum_j^{n^l}w_{h^{l}(i,j)}^l \xi^l(i,j) a_j^l\right)
\end{equation}
\begin{equation}
\delta_j^l=\left( \sum_{i=1}^{n^{l+1}} w_{h^{l}(i,j)}^l \xi^l(i,j)\delta_j^{l+1}\right)f^{'}(z_j^l)
\end{equation}
where $h^{l}(i,j)$ is the hash function associated with the weights in layer $l$ and $\xi^l(i,j): \mathbb{N}\rightarrow \pm 1$ is the second hash function independent of $h$ for sign function to remove the bias of hashed inner-products caused by collisions~\cite{weinberger2009feature}. $f^\prime \left (\cdot \right)$ represents the first derivative of activation function $f \left (\cdot \right)$, and $z_{j}^{l}$ is the result before activation function.
The weight transformation function will be modeled as $\varphi \left (W  \right )=W_{h}\bigodot \xi$ by introducing the two hash functions $h$ and $\xi$. Here $\bigodot$ denotes the elementary multiplication and the compression rate can be set by tuning $\xi$. Accordingly, augmented from the FGSM, we can derive the Hardware-oriented Fast Gradient Sign Method (HFGSM) dedicated to HashNet as:
\begin{equation}
X^{AE} = X + \epsilon sign(\nabla_XJ(X))\\
\end{equation}
where, the gradient can be calculated as:
\begin{equation}
\label{hashae}
%\resizebox{1\columnwidth}{!}{$
\begin{cases}
\nabla_XJ(X) = \sum_{i=1}^{n^{l=2}} w_{h^{l=1}(i,j)}^{l=1} \xi^{l=1}(i,j)\delta_j^{l=2}\\
\delta_j^l=\left( \sum_{i=1}^{n^{l+1}} w_{h^{l}(i,j)}^l \xi^l(i,j)\delta_j^{l+1} \right)f^{'}(z_j^l)
%\text{, } l=L-1,L-2,...,2
%\\
%\delta_j^{l}=\left( \nabla_{z_j^l}J(X) \right)f^{'}(z_j^l)\text{, } l=L
\end{cases}
%$}
\end{equation}

where $\epsilon$ is the amplitude coefficient of perturbations, $\nabla_XJ(X)$ is the gradient of loss function $J$ w.r.t. input $X$.

Similarly, the Hardware-oriented Jacobian-based Saliency Map Approach (HJSMA) for HashNet can be further developed with the same weight transformation but forward derivative gradient that can be obtained from the result of output layer. Thus an ``adversarial saliency map" that indicates \textit{the correlation between inputs and outputs} can be calculated from the gradient $\nabla_{x_i} F(x)$: 
% \begin{equation}
% \label{eq:sal}
% S(X,t)[i] = 
% \begin{cases}
% 0\text{ if }J_{it}(X) < 0\text{ or }\sum_{j\neq t}J_{ij}(X)>0\\
% J_{it}(X)\lvert\sum_{j\neq t}J_{ij}(X)\rvert\text{ otherwise}
% \end{cases}
% \end{equation}

% \begin{equation}
% \label{eq:_sal}
% S(X,T)[i] = 
% \begin{cases}
% 0\text{ if }\nabla_{X}T_i < 0\text{ or }\sum_{j\neq t\rvert j\in J, t \in T}\nabla_{X}J_i>0\\
% \nabla_{X}T_i\lvert\sum_{J\neq T}\nabla_{X}J_{i}\rvert\text{  otherwise}
% \end{cases}
% \end{equation}

\begin{equation}
\label{eq:sal}
S(X,t)[i] = 
\begin{cases}
0\text{ if }\nabla_{X_i} F_t(X) < 0\text{ or }\sum_{o\neq t}\nabla_{X_i}F_o(X)>0\\
\nabla_{X_i}F_t(X)\lvert\sum_{o\neq t}\nabla_{X_i}F_o(X)\rvert\text{  otherwise}
\end{cases}
\end{equation}

%\begin{equation}
%\label{hashae}
%\begin{cases}

%\nabla_XF(X) = \sum_{i=1}^{n^{l=2}} w_{h^{l=1}(i,j)}^{l=1} \xi^{l=1}(i,j)\delta_j^{l=2}\\
%\delta_j^l=\left( \sum_{i=1}^{n^{l+1}} w_{h^{l}(i,j)}^l \xi^l(i,j)\delta_j^{l+1} \right)f^{'}(z_j^l)
%\text{ if } l=	L-1,L-2,...,2
%\\
%delta_j^{l}=f^{'}(z_j^l)\text{ if } l=L
%\end{cases}
%\end{equation}

where each element of saliency map $S(X,t)[i]$ for a false target class $t$ is obtained based on the rule of rejecting input components with negative target derivative or an overall positive derivative on other classes $o$, otherwise accepting input components based on synthetic results of positive target derivative and all the other forward derivative components together. Therefore, only the input features corresponding to large values of $S(X,t)[i]$ in saliency map can be identified for adding adversarial perturbations, thus to efficiently mislead the classification result to a certain target.

\subsubsection{Attack Methodology}
To facilitate comprehensive adversarial attacks for the deep compressed DNN model, we develop
%Our attack design is following 
a synthesized attack methodology by integrating the derived HFGSM and HJSMA approaches. As Algorithm~\ref{alg} shows, 
%shows the attack methodology. 
%According to the adopted adversary generating approach, 
an upper-bound of the perturbation amplitude coefficient $\epsilon$ in HFGSM (see Eq.~\ref{hashae}) or the number of perturbation elements $i$ in HJSMA (see Eq.~\ref{eq:sal}) will be predefined to guarantee that the crafted adversarial perturbations can be maintained at an imperceptible level, which is more desirable in practical attacks. %In designed attack, 
A randomly selected original input-output pair ($X$, $Y$) will be recorded and compared with the adversarial input-output pair ($X^*$, $Y^*$). 
%and its original inference result $Y$ will be recorded for comparing with the generated adversary $X^*$ and adversarial inference result $Y^*$. 
%Straightforwardly, 
The adversarial example generating process will be terminated once a successful adversarial attack happens, i.e. $Y \neq Y^*$, otherwise $\epsilon$ or $i$ will be increased until reaching the respective upper-bound. The success rate of adversarial attacks will be adopted as a measurement in our following security analysis.

\begin{algorithm}[t]
\small
\caption{\label{alg}Adversarial Attack Methodology}
\DontPrintSemicolon
\tcp{$\mathcal{O}$ is the inference on target DNN model}
\tcp{$\mathcal{R}$ is the random selected inputs for a round of attack}
\tcp{$\epsilon$ is the amplitude coefficient of perturbation in HFGSM}
\tcp{$i$ is the number of perturbation elements in HJSMA}
\ForEach{$\vec{x}\in \mathcal{R}$}{
\tcp{get the original input X and inference result Y}
$D(X,Y) \leftarrow \left\{(\vec{x},\mathcal{O}(\vec{x}))|_{\vec{x}\in \mathcal{R}}\right\}$\;
\tcp{calculate the gradient s.t. input X}
$\nabla_XJ \leftarrow$ Equation~\ref{hashae}\;
\tcp{generating perturbation}
$\delta_{X_H} \leftarrow \text{HFGSM}(\nabla_XJ)\text{ or HJSMA}(\nabla_XJ)$\;
$X^* = X + \delta_{X_H}$\;
\tcp{perform inference using adversary as input}
$Y^* \leftarrow \mathcal{O}(X^*)$\;
\eIf{$Y = Y^*$}{
  \tcp{the adversarial attack is not success}
  \If{$\epsilon$ or $i$ \textless \text{ predefined upper-bound}}{
    increase $\epsilon$ in HFGSM (Equation~\ref{hashae}) or\;
    increase $i$ in HJSMA (Equation~\ref{eq:sal})\;
    GOTO: line 4\;
  }
}{
  adversarial success counter += 1\;
}
}

\end{algorithm}

\section{Security Analysis}
\label{SA}
We conduct the multi-factor adversarial attacks on the following tailored DNN model (i.e. 784-C64-C128-F512-10)
%\footnote{Network structure: input layer with 784 neurons -- 64 feature maps in first convolutional layer -- 128 feature maps in second convolutional layer -- fully connected layer with 512 neurons -- output layer with 10 neurons.} 
applied with HashNet model reshaping by following the proposed attack methodology.  %evaluated the attack effectiveness in Deep Learning system. 
A full MNIST database is adopted as our benchmark for a comprehensive analysis of attacking effectiveness in deep compressed/non-compressed deep learning systems.

%The multi-factor adversarial attacks are conducted by following the proposed methodology on  tailored DNN model (i.e. 784-C64-C128-F512-10)\footnote{input layer with 784 neurons -- 64 feature maps in first convolutional layer -- 128 feature maps in second convolutional layer -- fully connected layer with 512 neurons -- output layer with 10 neurons.} with HashNet model reshaping, to evaluated the attack effectiveness in Deep Learning system. A full MNIST database\footnote{60K training images and 10K testing images.} is adopted as benchmark.

\subsection{Effectiveness of Multi-factor Adversarial Attacks}
\begin{figure}[b]
\centering
\includegraphics[width = 1\columnwidth]{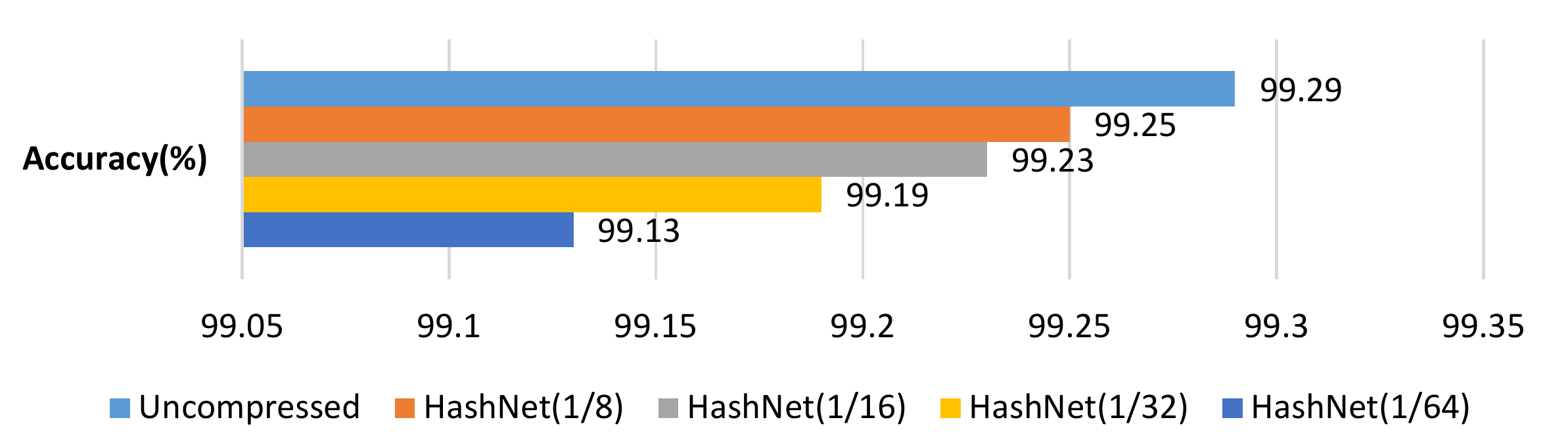}
\caption{Testing accuracy without adversarial perturbations.}
\label{fig:normalacc}
\end{figure}

%Before we perform the multi-factor adversarial attacks on HashNet, 
We first designed several hash compressed DNN models--HashNets with different compression rates (from $\frac{1}{8}$ to $\frac{1}{64}$) based on the aforementioned uncompressed DNN model. To make a fair adversarial attack analysis, our HashNets minimize the testing accuracy degradation (with normal input data without adversarial perturbations) introduced by weight compression. As shown in Fig.~\ref{fig:normalacc}, the testing accuracy on HashNets is only slightly decreased as the compression rate increases (i.e. 99.25\% at rate $\frac{1}{8}$ v.s. 99.13\% at $\frac{1}{64}$) but still very close to the uncompressed model (99.29\%).

%make a fair adversarial attack analysis, our HashNet should maintain the same 
%we further evaluated the testing accuracy without adversarial perturbations. 

\begin{figure}[t]
\centering
\includegraphics[width = 1\columnwidth]{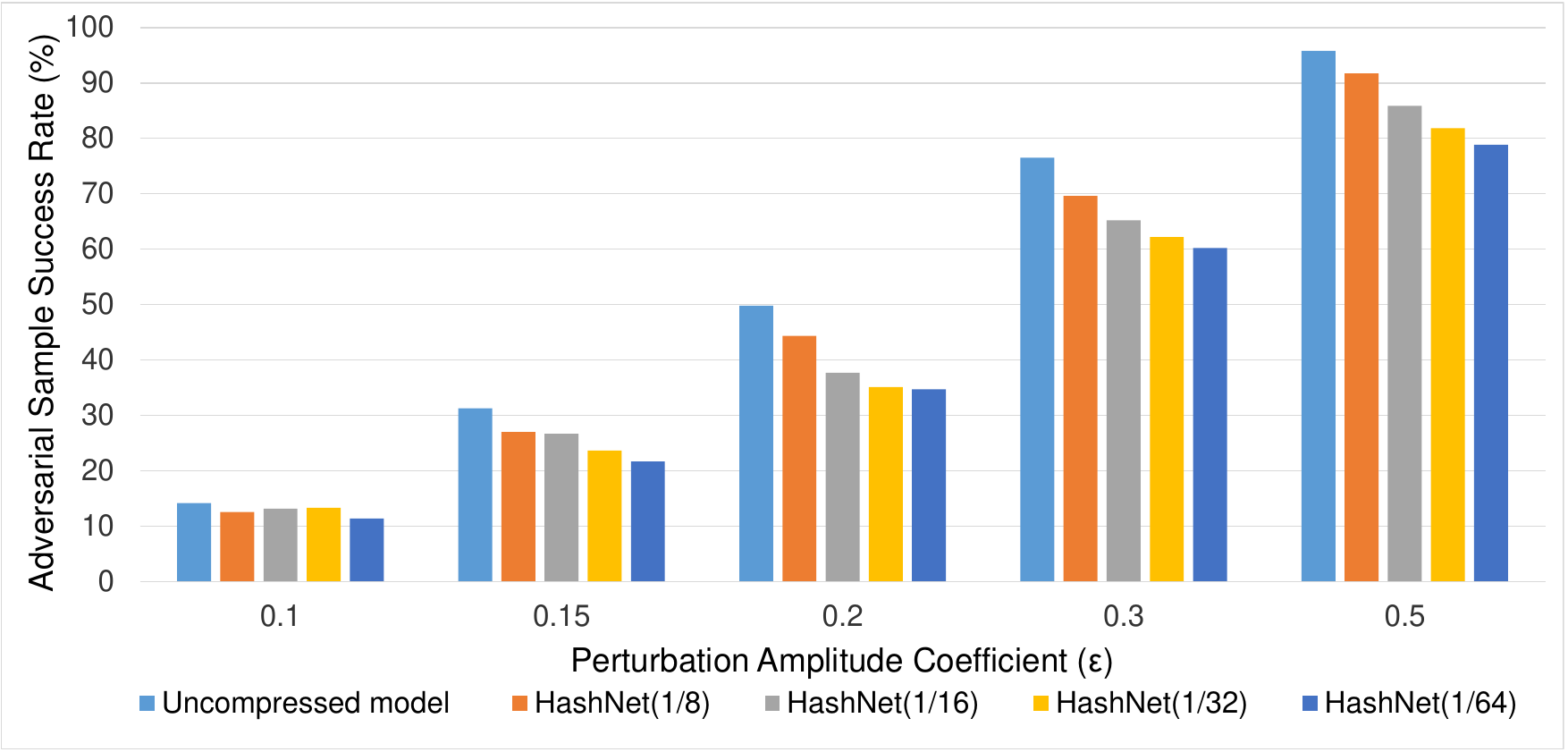}
\caption{Success rate of multi-factor adversarial attacks with HFGSM approach.}
\label{fig:FGSM_success_rate}
\end{figure}

\begin{figure}[t]
\centering
\includegraphics[width = 0.93\columnwidth]{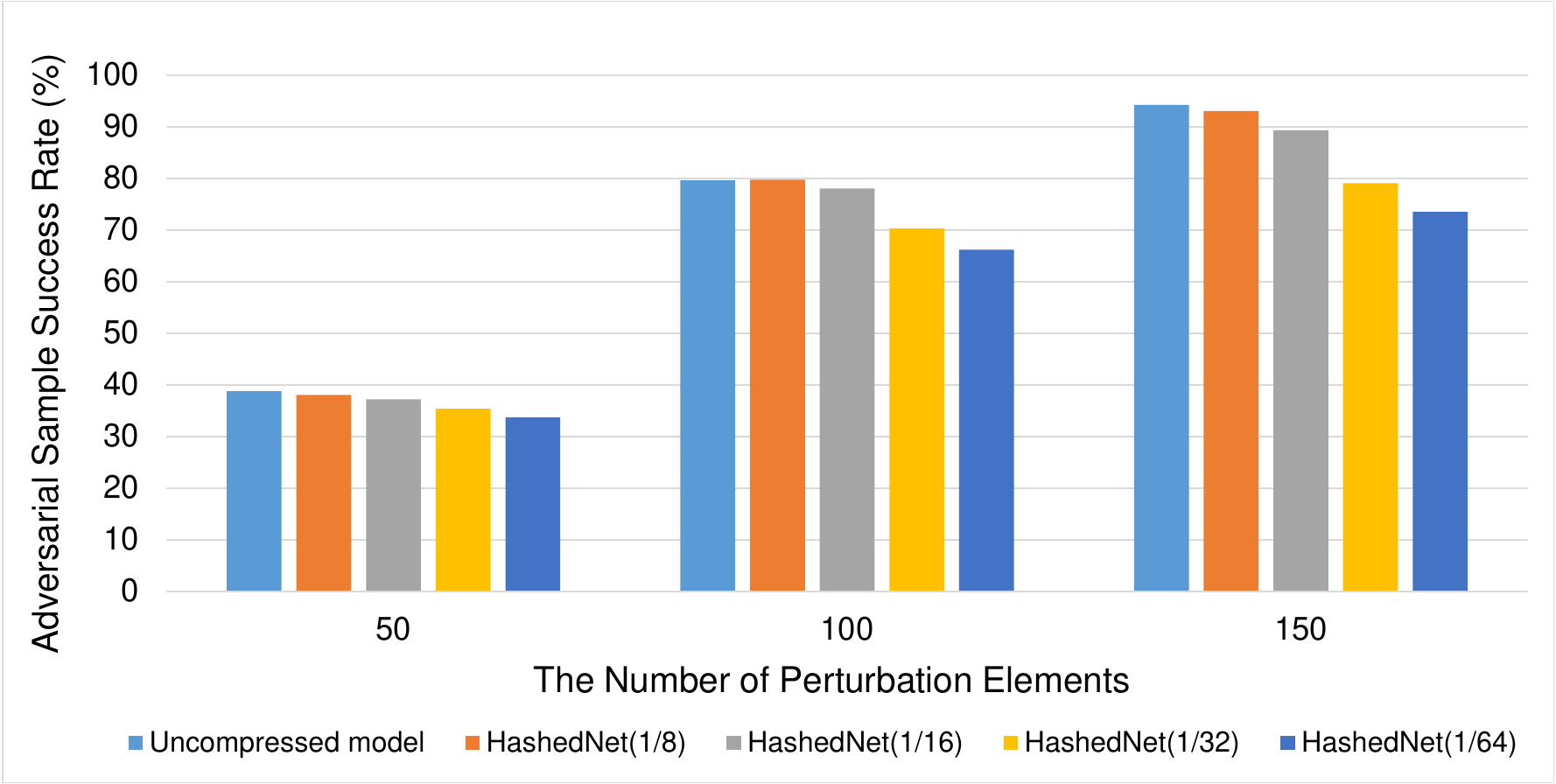}
\caption{Success rate of multi-factor adversarial attacks with HJSMA approach.}
\label{fig:jsma}
\end{figure}

Fig.~\ref{fig:FGSM_success_rate} shows the success rate of multi-factor adversarial attacks implemented with HFGSM method at various compression rates (i.e. HashNet($\frac{1}{8}$) $\rightarrow$ HashNet($\frac{1}{64}$) over different perturbation amplitude coefficients. %$\epsilon = 0.1 \rightarrow 0.5$
For comparison purpose, the results of the uncompressed DNN model--the common basis of different HashNets, under the original single-factor based FGSM attacks are also presented.
%Fig.~\ref{fig:FGSM_success_rate} shows the effectiveness of multi-factor adversarial attacks measured by the attack success rate. Several designs with different compression rate, i.e. HashNet($\frac{1}{8}$) $\rightarrow$ HashNet($\frac{1}{64}$), are implemented with HFGSM to compare with the uncompressed DNN model (in same structure) and original FGSM approach.
%As shown in Fig.~\ref{fig:FGSM_success_rate}, 
As expected, the attack success rates of both uncompressed DNN model and various compressed models are increased monotonically %monotonously 
along with the growing perturbation amplitude coefficient, i.e. $\epsilon = 0.1 \rightarrow 0.5$. 
%as expected. 
This is because the attacking capability of crafted adversarial examples can be significantly enhanced by larger input perturbations (see $\epsilon=0.5$) for all DNN models regardless of the model reshaping.
%Larger amplitude of perturbation denotes enhanced adversarial strength, thus higher attack success rate. 
However, for each individual $\epsilon$, the attack success rates of any HashNet models are always lower than that of uncompressed model. Moreover, the higher the compression rate is, the lower the attack success rate will be at each $\epsilon$. We also conduct the same set of experiments under HJSMA based adversarial attacks. Again, our results in Fig.~\ref{fig:jsma} demonstrate the similar trend at different combinations of compression rate and the number of perturbation elements, i.e. the attack success rates are decreased when increasing compression rate on HashNet at each selected number of perturbation elements. Surprisingly, \textit{these results indicate that the hash compressed DNN model, which have significantly reduced number of model parameters for affordable hardware implementation (see Fig.~\ref{fig:Hash}), exhibits better resistance to adversarial attacks than that of its uncompressed or less compressed version. This is in contrast to the empirical intuition that the more compressed DNN models should be more susceptible to the input perturbations.} 
%and show reductions along with the increasing compression rates. 
%This result is out of our expectations. 
%To verify this problem, adversarial attacks with HJSMA approach are evaluated as well with the same methodology and configurations. As shown in Fig.~\ref{fig:jsma}, for each setting of perturbation elements, the success rates of adversarial attacks are decreased when increasing compression rate on HashNet, presenting the same phenomenon as shown in Fig.~\ref{fig:FGSM_success_rate}.
%These evaluations show highly compressed model has better resistance to adversarial attacks. 

%Empirically, models with higher compression rate maintain the reduced number of weights due to the intensified model reshaping and weights transformation (w.f.t. Fig.~\ref{fig:Hash}), indicating compromised functionality and stability, thus more susceptible to the input perturbations. 

%To verify the stability of compressed model, we further evaluated the testing accuracy without adversarial perturbations. As shown in Fig.~\ref{fig:normalacc}, the testing accuracy on HashNet model is slightly decreased from 99.25\% to 99.13\% by increasing compression rate from $\frac{1}{8}$ to $\frac{1}{64}$ and all of them are lower then the uncompressed model (99.29\%), 
%indicates minor degradations on neural network stabilities. 

Since the compressed DNN models maintain the similar level of the stability (or testing accuracy in Fig.~\ref{fig:normalacc}) as that of uncompressed model, a reasonable explanation for the attack success rate reduction is that the destructiveness of crafted adversaries may be alleviated in HashNets when compare with those generated in uncompressed DNN model. \textit{That is being said, the effectiveness of multi-factor adversary attacks depends on the perturbation amplitude coefficient $\epsilon$ in HFGSM (or the number of perturbation elements $i$ in HJSMA) and the compression rate,} as we shall discuss in the following section.

%\subsection{Semi-analytical analysis of Compression rate, Weight and Gradient}
\subsection{Theoretical Analysis of Adversarial Attacks on Hashed DNNs}

\begin{figure}[b]
\centering
\includegraphics[width = 1\columnwidth]{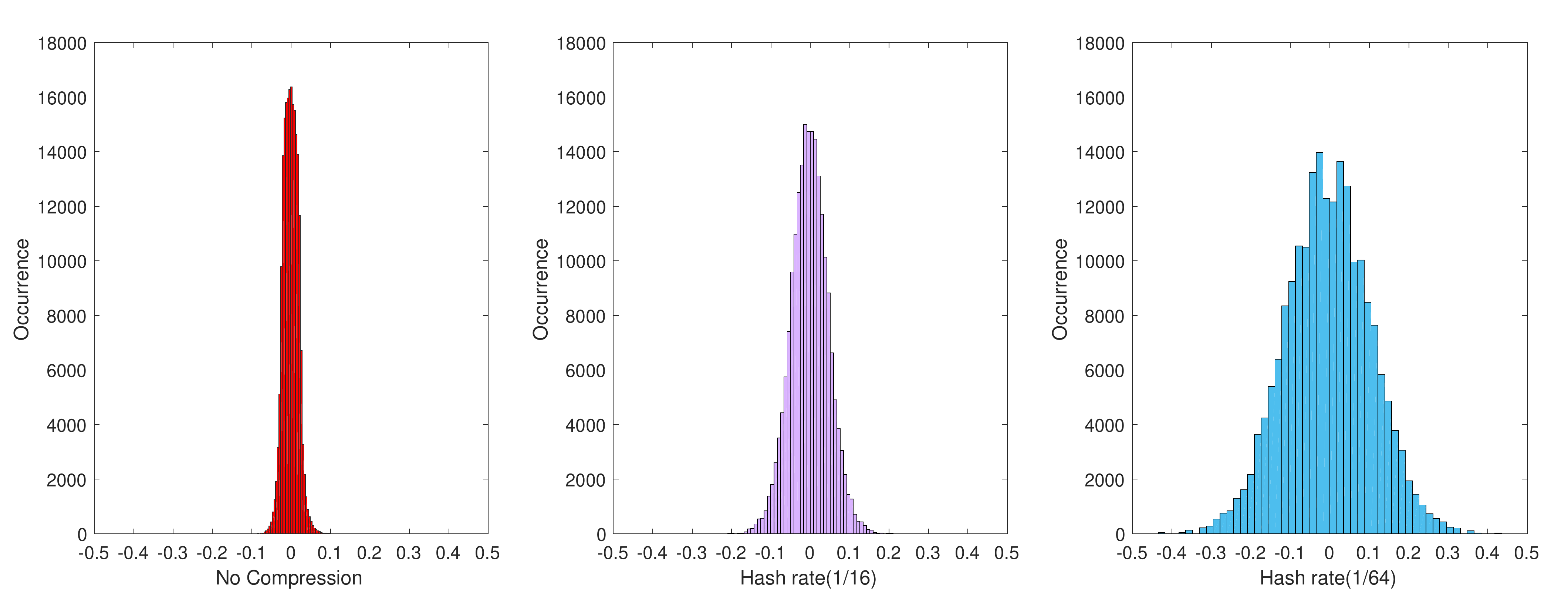}
\caption{The weight distributions for uncompressed DNN and two HashNets.}
\label{fig:weight}
\end{figure}

To validate our hypothesis and deeply understand the relationship between adversary and model reshaping, we characterize the two critical components for adversary example generating in compressed models: weight and gradient amplitude under various compression rates. 
%The transformation of weights and gradients in compressed model has been analyzed in this work. Notice that the weight and gradient are two fundamental components in adversary generating. 
Fig.~\ref{fig:weight} compares the distributions of weights for uncompressed and two compressed DNNs--HashNet ($\frac{1}{16}$) and HashNet ($\frac{1}{64}$).
%shows the histogram of weights in . Three designs with different compression rate, i.e. no compression (uncompressed model), $\frac{1}{16}$ and $\frac{1}{64}$, are selected to demonstrate the weights transformation in model reshaping. As shown in 
As Fig.~\ref{fig:weight} shows, the model with a higher compression rate yields a larger range of weights (i.e. $\sim2\times$ and $\sim4\times$ in HashNet ($\frac{1}{16}$) and HashNet ($\frac{1}{64}$) w.r.t. uncompressed model). Given significantly decreased number of unique weights (or increased compression rate) introduced by hash-based weight sharing mechanism, the weight distribution in compressed DNN model shall be much broader since such model has to re-balance the activations through enlarged weights during training to achieve an accuracy close to that of uncompressed model. However, such weight transformation can directly impact the gradient, thus the strength of generated adversaries.
%can lead to enlarged weights 
%a much broader weight distribution, thus 
%thus enlarged activation in neural network processing. 

Without loss of generality, we use the output layer with softmax activation to roughly explain the underlying principle. 
The final activation of output layer can be calculated through the following \textit{Softmax} function:
%Consider the neural network processing in the output layer, in which the final activation will be calculated through the \textit{Softmax} function to make the decision on inference result:
\begin{equation}
F(z_j)=\frac{e^{z_j}}{\sum_{k=1}^n e^{z_k}}
\end{equation}
where the input $z_j$ of \textit{Softmax} function can be expressed as:
\begin{equation}
z_j=\sum_{i=1}^n w_{ji}x_i
\end{equation}
Note that we omit the bias because it can be included in weight by adding an additional connection with weight as the bias and a constant input 1. Since the \textit{Softmax} function increases monotonically as the input $z_j$ grows, \textbf{the enlarged weights in highly compressed models can possibly augment the desired activations but suppress the others, thus a possible stronger confidence for the final decision.} 

%The explain of weights enlargement is still uncertain, probably due to the decreased number of weights, neural network has to re-balance the activations through enlarged weights in its training. 

\begin{figure}[t]
\centering
\includegraphics[width = 0.86\columnwidth]{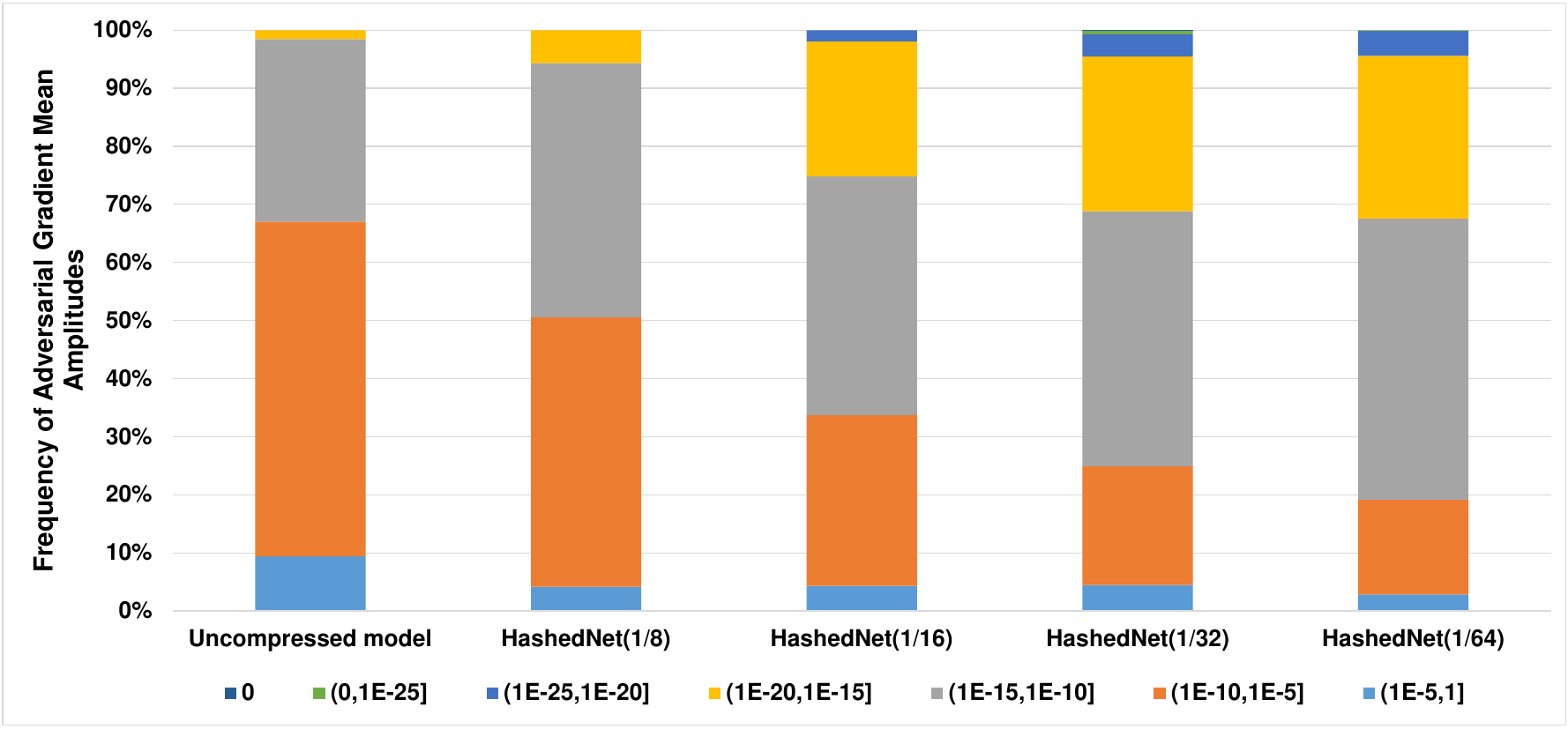}
\caption{Absolute gradient amplitude with different compression rate.}
\label{fig:gradient}
\end{figure}

%However, the consequence is the enlarged weights will directly impact the gradient, thus significantly affect the generated adversaries. 
%Notice that 
If we use FGSM based adversarial example generating algorithm as an example, the cross-entropy loss function and its gradient w.r.t. input can be obtained as:
\begin{equation}
J(x_i,t_j) = -\sum_{j=1}^n t_j \log F(z_j) 
\end{equation}
%therefore, the gradient from output layer is:
\begin{equation}
\nabla_{x_i}J(x_i,t_j) = \sum_{j=1}^n w_{ji}(F(z_j)-t_j)
\end{equation}
where $x_i$ is the $i^{th}$ input and $t_j$ is the target for $j^{th}$ class. Consider the $F(z_j)$ is an exponential function of $w_{ji}$, the absolute gradient amplitude will be dominated by term $(F(z_j)-t_j)$. \textbf{With the enlarged weights in compressed models, the activation $F(z_j)$ may be closer to $t_j$, 
% \begin{equation}
% \lim_{|w_{ji}|\rightarrow\infty}(F(z_j)-t_j) = 0 
% \Rightarrow \nabla_{x_i}J(x_i,t_j) \rightarrow 0
% \end{equation}
thus a possible reduced gradient and perturbation amplitude, meaning alleviated adversarial severity.} Fig.~\ref{fig:gradient} shows the distributions of absolute value of mean gradient over uncompressed and compression DNNs with different compression rates. The proportion of large gradients ($10^{-10} \sim 1$) is reduced from $\sim68\%$ (uncompressed model) to $\sim19\%$ (HashNet($\frac{1}{64}$)) as compression rate grows, while that of small gradients ($10^{-25} \sim 10^{-15}$) is increased from $\sim1\%$ to $\sim32\%$, which is in excellent agreement with our theoretical analysis and validates the degraded attack capability of compressed DNNs compared with the uncompressed version.

\section{Mitigation Approach}

\begin{table}[b]
\centering
\begin{tabular}{|l|c|c|c|c|}
\hline
Candidate Models     & DNN1     & DNN2     & DNN3     & VGG-16 \\ \hline
Relu Convolutional   & 4 layers & 6 layers & 9 layers & 13 layers      \\ \hline
Relu Fully Connected & 2 layers & 2 layers & 2 layers & 3 layers      \\ \hline
Max Pooling          & 2 layers & 3 layers & 3 layers & 5 layers      \\ \hline
\end{tabular}
\caption{\label{tab:modelcan}Architectures of selected neural network candidates.}
\end{table}

In our security analysis, 
%the relationship between weights, gradients and adversary is explored thoroughly. 
we show that the magnitude of weights in DNNs becomes a new factor that can significantly impact the the severity of adversarial attacks. Hash-based weight compression enlarges the magnitude of weights, thus to prevent the generating of stronger adversarial examples. However, its effectiveness is very limited, e.g. $\le30\%$ success rate reduction at any perturbation amplitude coefficient in Fig.~\ref{fig:FGSM_success_rate}, because the weight enlargement, introduced by non-linear weight transformation, can only be guaranteed at a certain probability (see Fig.\ref{fig:weight}). Inspired by this observation, a novel mitigation technique named \textit{Gradient Inhibition} is further proposed to effectively mitigate the adversarial attacks.

%the weight compression cannot  
%due to the uncertainty between weights compression and enlargement,

%the reduction of adversarial severity is still incontrollable and only effect on Deep Learning System.

\subsection{Gradient Inhibition method}
%Base on our explorations, 
Our proposed \textit{Gradient Inhibition} intends to control the weights linearly with enlarged magnitude guarantee for each weight:
\begin{equation}
w = w + \tau * sign (w)
\end{equation}
where $\tau$ is the inhibition coefficient. Different levels of weight enlargement can be achieved by a fine-grained control parameter $\tau$ for both positive and negative weights, thus to minimize the gradient needed for adversarial perturbations generating and effectively mitigate or even eliminate the threats of adversarial attacks for DNNs.

%has been precisely leveraged in Gradient Inhibition method to efficiently minimize the gradients in neural network models, preventing the generating of adversarial perturbations, thus 

Another advantage of \textit{Gradient Inhibition} method is its low implementation cost applicable to both software or hardware-oriented compressed DNN models.
%and hardware-favorable computation. 
%Besides, through the adjustable inhibition coefficient, magnitude of weights can be intentionally changed towards their expectations, thus efficient suppression on gradients and adversarial perturbations. 
\textit{Gradient Inhibition} can be applied at any layer after the training process. Our practice is to deploy this method at the layers close to the output layer (i.e. the last fully connected layer) for higher attack rate reduction but lowest accuracy loss due to the usually moderate number of weights and strongest impacts on decision making.

\subsection{Evaluation of Gradient Inhibition}
%In this section, we evaluate the performance of proposed Gradient Inhibition mitigation technique.

\begin{figure}[t]
\centering
\includegraphics[width=0.86\columnwidth]{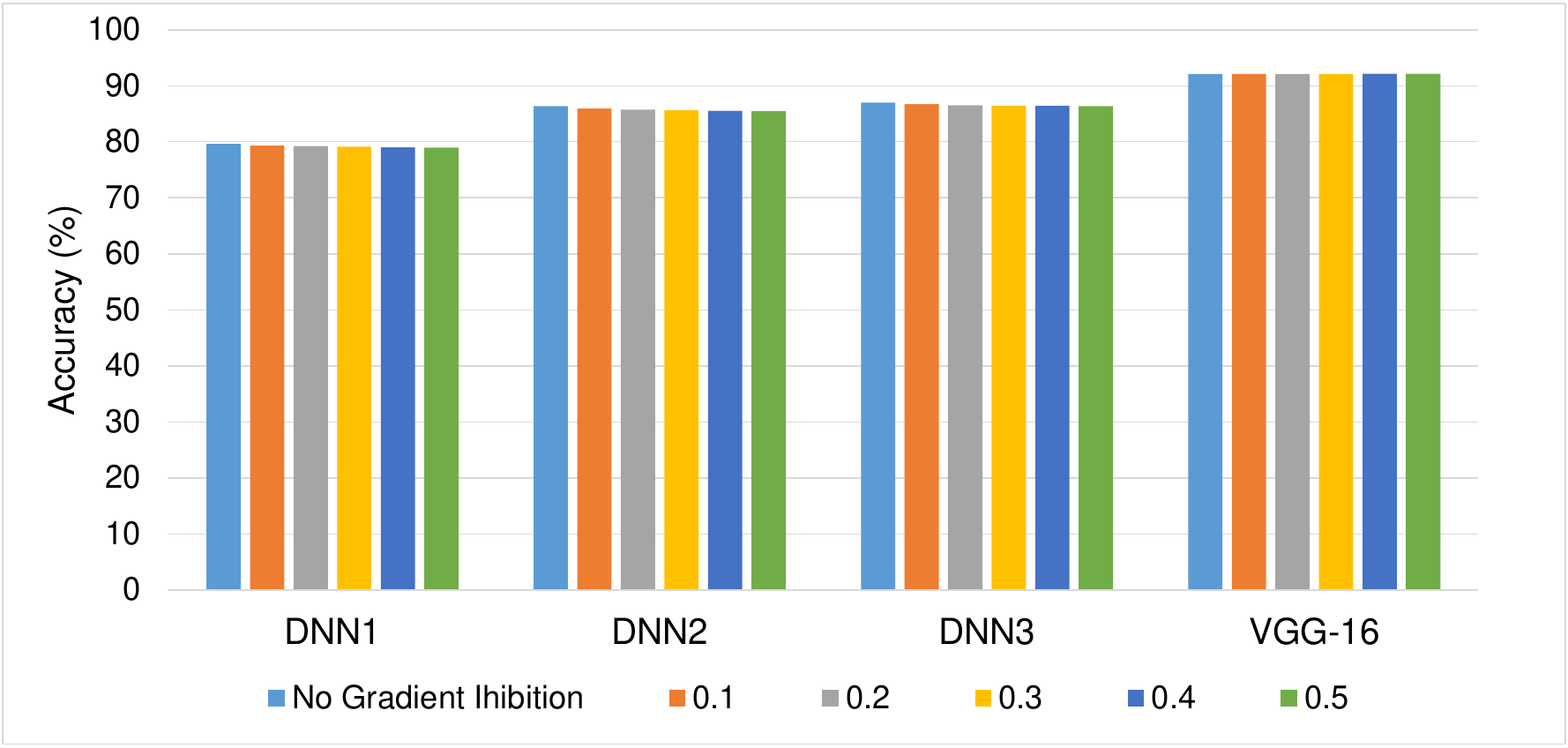}
\caption{\label{fig:accuracy_cifar10_new_method}Inference accuracy of CIFAR-10 with Gradient Inhibition}
\end{figure}

\begin{figure}[b]
\centering
\includegraphics[width=0.88\columnwidth]{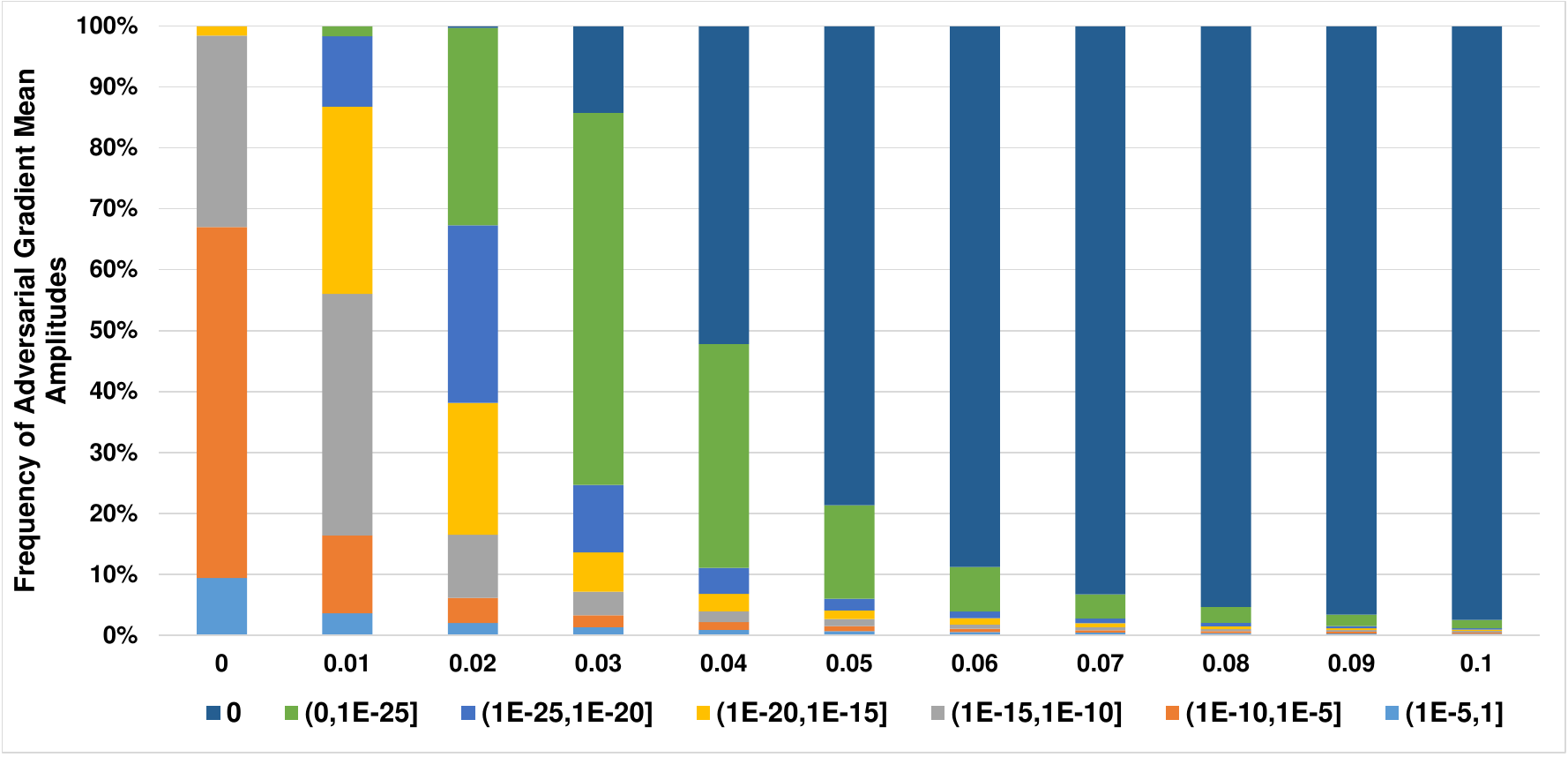}
\caption{\label{fig:amplitude_new}
Absolute gradient amplitude of uncompressed DNN at various inhibition coefficients}
%(Gradient Inhibition method on uncompressed model)}
\end{figure}

\subsubsection{Experiment Setup}
Various HashNets and MNIST benchmark~\cite{lecun1998mnist}, which are used in section~\ref{SA}, are adopted in our experiment to evaluate efficiency of \textit{Gradient Inhibition}. Additionally, the CIFAR-10 database~\cite{krizhevsky2009learning} is selected as a new benchmark in our evaluation, including 60K 32$\times$32 color images in 10 classes, 50K for training and 10K for testing. As shown in Table.~\ref{tab:modelcan}, four representative DNN models with different architectures, including state-of-art VGG-16~\cite{simonyan2014very}, are chosen to verify the feasibility and scalability of Gradient Inhibition across various types of DNN models. We assume the adversarial examples are generated through the FGSM and HFGSM for uncompressed and compressed models, respectively.

\subsubsection{Inference Accuracy}
An effective mitigate technique against adversarial attacks should not impact the functionality of the DNN models integrated with mitigate techniques. Before we evaluate the effectiveness, we first verify the inference accuracy changes introduced by \textit{Gradient Inhibition}. As shown in ~\ref{fig:accuracy_cifar10_new_method}, the inference accuracy on CIFAR-10 database for each DNN model implemented with \textit{Gradient Inhibition} is always at the same level as that of its corresponding model without such technique at different inhibition coefficients. We also find the similar accuracy trend in Hash compressed DNNs with different compression rates for the MNIST dataset. Note that the adopted inhibition coefficient $\tau = 0.1 \rightarrow 0.5$ can introduce flexible weight adjustments, i.e. $\pm0.1 \rightarrow \pm0.5$, with very minor accuracy change.
%Our proposed %\textit{Gradient Inhibition} method will not cause any side effect.
%Even larger than the original weights, perfectly demonstrating our proposed Gradient Inhibition method will not cause any side effect on the host neural network model.

\subsubsection{Gradient Inhibition Efficiency}

Fig.~\ref{fig:amplitude_new} shows the statistics of suppressed gradients across various inhibition coefficients for an uncompressed DNN model testing the MNIST dataset. %with proposed Gradient Inhibition method. 
%To fairly compare the results with the intrinsic gradient suppression in HashNet with model compression (see Fig.~\ref{fig:gradient}), this evaluation is conducted on the same uncompressed model. 
As shown in Fig.~\ref{fig:amplitude_new}, even with a very small adjustment on original weights, i.e. inhibition coefficient $\tau$ = 0.01, the gradient amplitude can be much lower than the one generated on HashNet($\frac{1}{64}$) in Fig.~\ref{fig:gradient}, which is the best case in compressed DNN models. Note that in HashNet($\frac{1}{64}$), the range of weights has been enlarged from $\pm0.1$ to $\pm0.4$ (see Fig.~\ref{fig:weight}), which is far exceed that of $\pm0.01$ in Gradient Inhibition. Therefore, our proposed method can significantly suppress the gradients with much lighter weight transformation. %thus better efficiency. 
Moreover, as shown in Fig.~\ref{fig:amplitude_new}, most of gradients are approaching to ``0" along with the increased inhibition coefficient, indicating the possible elimination of adversarial perturbations, thus to prevent the adversarial attacks remarkably.

%  The accuracy variation of DNN model and hashNet DLS model trained on MNIST dataset for the set of resistance coefficient on are shown in Figure ~\ref{fig:accuracy_variation_mnist_new_method}.We can observe that gradient inhibition method has very little effect on accuracy, whether it is DNN model or hashedNet DSL models trained on MNIST dataset.For instance,the accuracy of the models is dropped by less than 0.43\% on average for all inhibition coefficient.The least drop in accuracy is the DNN model(dropped by 0.15\%), and the most drop is the hashNet DSL model of compression factor 1/16(dropped by 0.86\%).
 
%  The accuracy variation of 4 different DNN models trained on CIFAR10 dataset for the set of inhibition coefficient on are shown in figure ~\ref{fig:accuracy_variation_cifar10_new_method}.we can observe that the accuracy of the models is dropped by less than 0.53\% on average for all inhibition coefficient.Interestingly, the  accuracy of VGG-16 model didn't drop for almost inhibition coefficient,and the success rate of adversarial attack decreased obviously(from 87.75\% to 7.84\%), which shows that our gradient inhibition method can be widely applied to very deep DNN.

\subsubsection{Mitigation Measures}
\begin{figure*}[t]
\centering
\begin{subfigure}{0.45\textwidth}
\includegraphics[width=1\columnwidth]{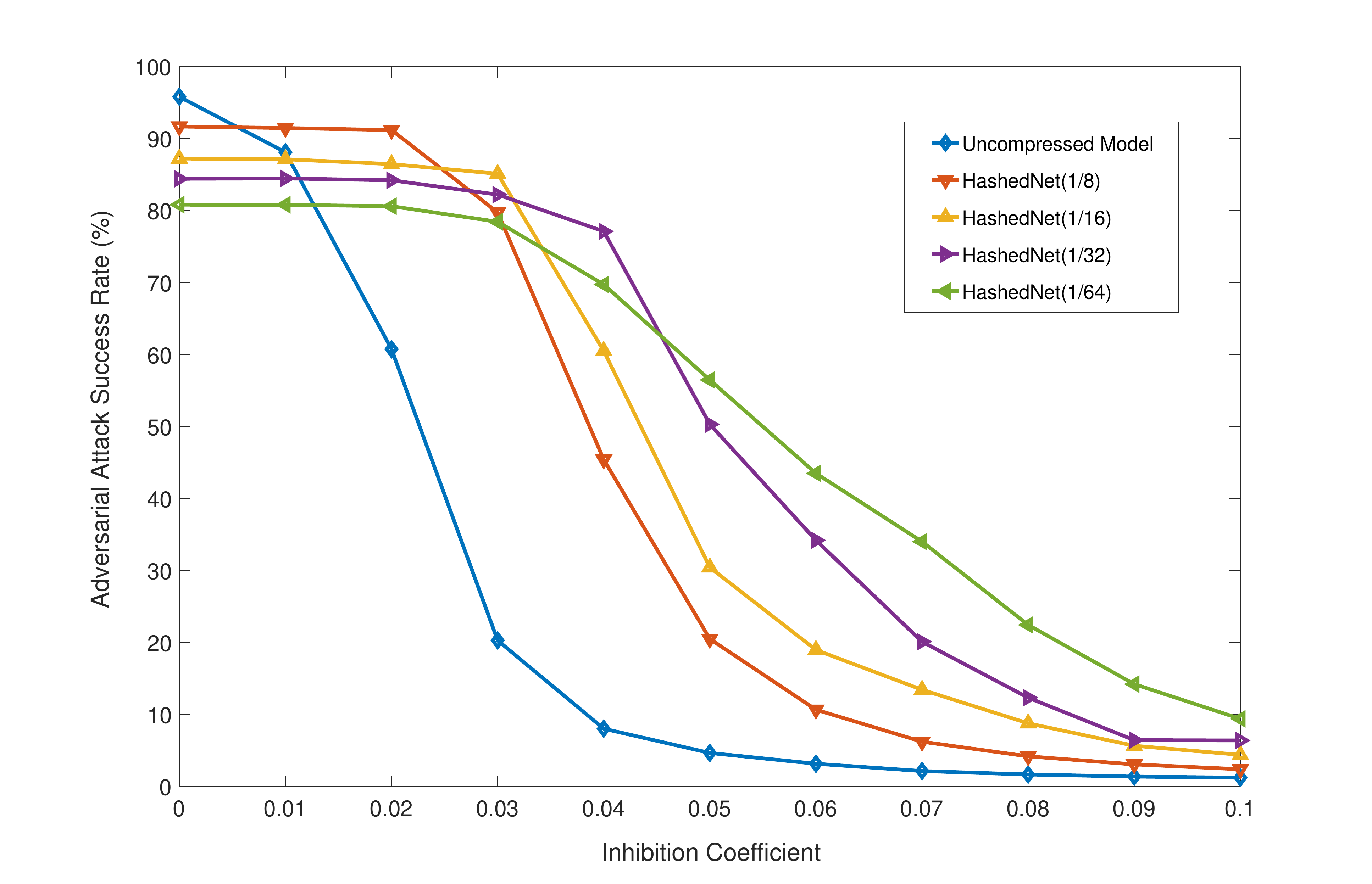}
\caption{\label{fig:AE_success_rate_new_method}HashNet with model compression -- MNIST}
\end{subfigure}
\begin{subfigure}{0.45\textwidth}
\includegraphics[width=1\columnwidth]{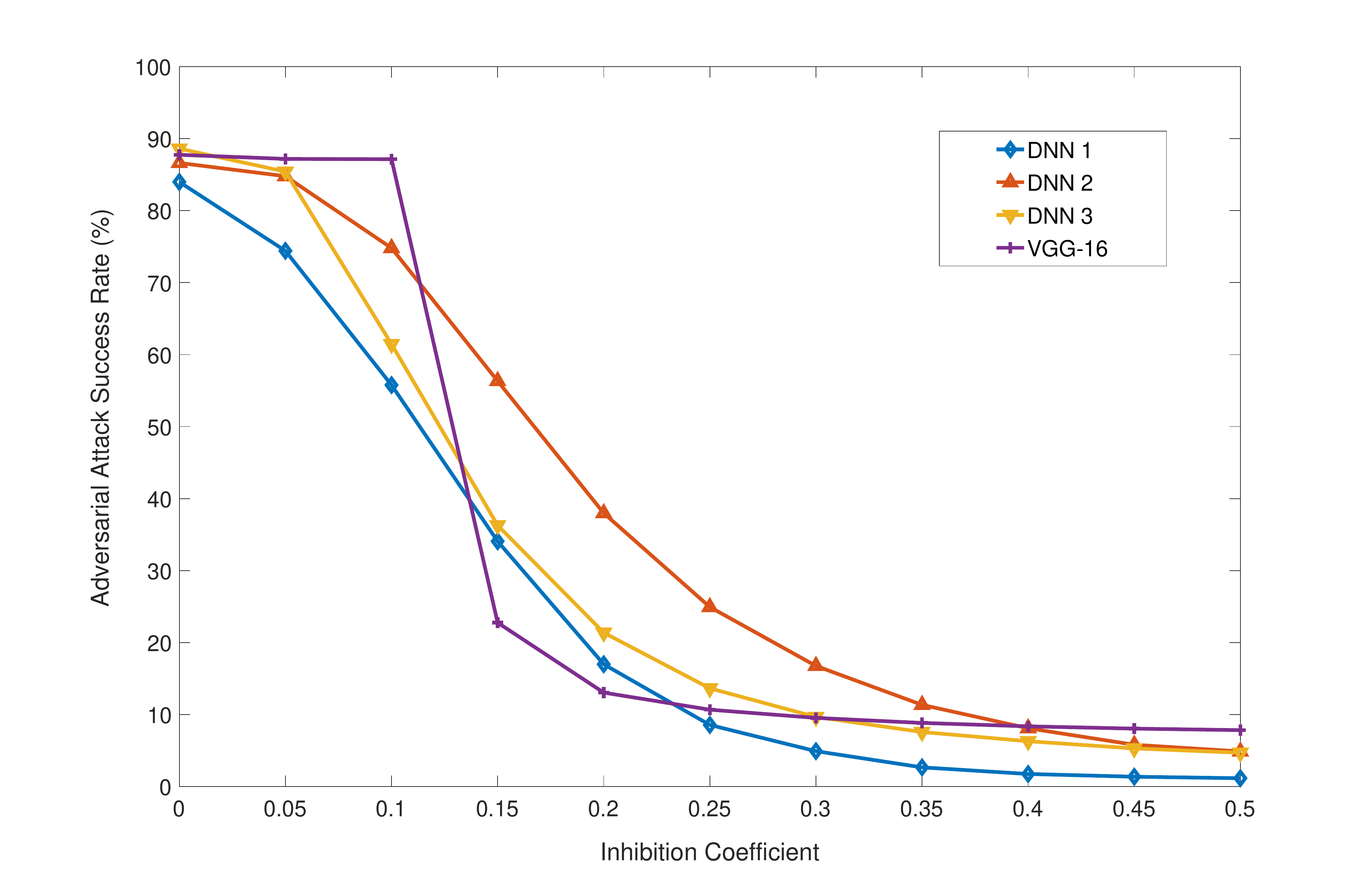}
\caption{\label{fig:AE_success_rate_new_method_cifar10}DNNs -- CIFAR-10}
\end{subfigure}
\caption{\label{fig:sucrate_gi}Success rate of adversarial attacks with Gradient Inhibition mitigate technique.}
\end{figure*}

Adversarial attacks are conducted by following the proposed attack methodology, on both DNN and compressed HashNet models with the Gradient Inhibition method. 
%Gradient Inhibition method is applied on host models with different inhibition coefficient configuration before we launch the attack. The success rate of adversarial attacks is adopted to measure the effectiveness of Gradient Inhibition method. 
%The lower attack success rate indicates the better mitigation.
Fig.~\ref{fig:sucrate_gi} (a) and (b) show the success rates of adversarial attacks under Gradient Inhibition over HashNets (for MNIST) and four DNNs (for CIFAR-10), respectively. As Fig.~\ref{fig:AE_success_rate_new_method} shows, the average success rate of adversarial attacks (HashNets, perturbations crafted through HFGSM with $\epsilon = 0.5$) can be reduced from 87.99\% to 4.77\% by increasing the inhibition coefficient $\tau$ from 0 to 0.1. Specifically, the uncompressed model presents the best efficiency ($95.81\% \rightarrow 1.24\%$) while all compressed HashNets exhibit some resistance to Gradient Inhibition and eventually reduce the adversarial success rate to less then 10\% at all selected compression rates. Fig.~\ref{fig:AE_success_rate_new_method_cifar10} evaluate the efficiency of proposed Gradient Inhibition on DNNs with CIFAR-10 database. The average success rate is dropped from 86.74\% to 4.64\% across various DNNs, demonstrating effective mitigations for adversarial attacks.

\section{Conclusion}

%Deep Neural Networks (DNNs) 
%and 
%Deep Learning System (DLS) 
%exposure 
%exposes themselves to serious security challenges of the emerging adversarial attacks. 
The emerging adversarial attacks leave the prevalent hardware accelerated Deep Neural Networks (DNNs) exposed to hackers. However, existing DNN security researches solely focus on the input perturbations but neglect the impacts of model-reshaping essential for DNN hardware deployment. In this work, the multi-factor adversarial attack problem is for the first time modeled and studied through extensive experimental and theoretical analysis. Based on the explorations of model-reshaping and adversarial examples generating, a novel mitigation technique -- ``Gradient Inhibition" is further proposed to effectively alleviate the severity of adversarial attacks for various DNNs. Our simulations demonstrate that ``Gradient Inhibition" can significantly reduce the success rate of adversarial attacks while maintaining the desired inference accuracy without additional trainings. We hope that our results enable the community to examine the emerging security issues of hardware-oriented DNNs.

\bibliographystyle{IEEEtran}
\bibliography{refs}

% {\onecolumn
% \input{outline}
% %\input{experiment}
% \input{fu}
% }

\end{document}